\documentclass{article} 
\usepackage{iclr2016_conference,times}
\usepackage{hyperref}
\usepackage{url}
\usepackage[utf8]{inputenc} 

\usepackage{amsmath,amsfonts,amssymb}
\usepackage{graphicx}
\usepackage{booktabs}
\usepackage{caption}

\DeclareMathOperator{\softmax}{softmax}
\DeclareMathOperator{\sig}{sigmoid}

\title{A Cheap Linear Attention Mechanism with Fast Lookups and Fixed-Size \mbox{Representations}}

\author{Alexandre de Brébisson \\
MILA, University of Montréal\\
\texttt{alexandre.de.brebisson@umontreal.ca}
\And
Pascal Vincent \\
MILA, University of Montréal\thanks{ and CIFAR}\\
\texttt{vincentp@iro.umontreal.ca}
}

\iclrfinalcopy 

\begin{document}

\maketitle

\begin{abstract}

The softmax content-based attention mechanism has proven to be very beneficial in many applications of recurrent neural networks. Nevertheless it suffers from two major computational limitations. First, its computations for an attention lookup scale linearly in the size of the attended sequence. Second, it does not encode the sequence into a fixed-size representation but instead requires to memorize all the hidden states. These two limitations restrict the use of the softmax attention mechanism to relatively small-scale applications with short sequences and few lookups per sequence. In this work we introduce a family of linear attention mechanisms designed to overcome the two limitations listed above. We show that removing the softmax non-linearity from the traditional attention formulation yields constant-time attention lookups and fixed-size representations of the attended sequences. These properties make these linear attention mechanisms particularly suitable for large-scale applications with extreme query loads, real-time requirements and memory constraints. Early experiments on a question answering task show that these linear mechanisms yield significantly better accuracy results than no attention, but obviously worse than their softmax alternative.

%
%
%
%
%

\end{abstract}

%

\section{Introduction}

Many large-scale applications, in particular among information retrieval tasks, require efficient algorithms to compress documents and query them. For example, at test time, systems may have to process millions of queries simultaneously and in real-time. The content-based attention mechanism~\citep{Bahdanau-et-al-ICLR2015-small} is a recently introduced architecture that allows the system to focus on particular parts of the document depending on the query. It has proven to be very beneficial in many applications of deep learning but its expensive computations often prevent it from being used in large-scale applications. In this work we introduce a family of linear attention mechanisms that overcome these limitations and still offer to some extent the benefits of the traditional attention mechanism.

\textbf{Notations}: Let $D$ represent a document sequence of $n$ tokens and let us consider $m$ queries on this document. Let $Q$ represent one of these queries, which is encoded into a column vector representation $q$ (for example the last state of a recurrent neural network). The document $D$ is processed with a recurrent neural network, which, at each timestep $t$, computes a hidden state $h_{(t)}$ of size $k$. Let $H$ be the $n \times k$ matrix composed of all the hidden states of the document $D$ stacked vertically, i.e. whose row $H_{t.} = h_{(t)}$. 

\newpage

\section{Classic softmax attention mechanism}

\subsection{Definition and complexity}

In this work, we consider the following form of softmax attention mechanism\footnote{Note that this form is found in memory networks~\citep{NIPS2015_5846} but other forms are common (all with similar complexities and memory requirements), in particular the one introduced by \citet{Bahdanau-et-al-ICLR2015-small}. We present this particular form because it is the most similar to the cheap mechanism that we introduce in the next section.}, which computes a representation $R(D,Q)$ of the document $D$ conditioned on the question $Q$:

\[
R(D,Q) = H^T \softmax(H q),
\]

where $H q$ represents the inner products of $q$ with all the hidden states of the document $D$. The softmax then converts these inner products into probabilities that are used to compute a weighted sum of the hidden states stacked in~$H$.

This mechanism involves matrix multiplications which result in an overall $O(n k^2)$ complexity for a single query $Q$ lookup. If, instead of considering a single query, we would like to process $m$ queries, the complexity would be $\boldsymbol{O(m n k^2)}$. If $n$ or $m$ are very large, this complexity is prohibitive and restricts the scale of the potential applications.

Furthermore, the classic softmax attention mechanism does not allow to store a fixed-size representation of the document $D$. Instead, all of the hidden states of the network have to be stored, resulting in a variable-size representation that requires $O(n k)$ memory space. This can also prohibitive when $n$ is large.

\subsection{Applications of the softmax attention mechanisms and limitations}

In this section, we describe a few use cases of the softmax attention mechanism and how its computational cost may limit the scale of its applications.

\begin{itemize}
	\item In machine translation~\citep{Bahdanau-et-al-ICLR2015-small}, the document $D$ would be the source sentence that has to be translated and which is composed of $n$ words. The translated sentence is generated iteratively and at each new timestep, an attention lookup $Q$ is performed. The number of words of the translated sequence is $m$, which corresponds to the number of required attention lookups. Thus, for each new generated word, a new $O(n)$ attention lookup is performed. This may significantly slow down the translation of long sentences (large $n$ and large $m$) and prevent real-time translations.
	\item In question answering~\citep{nips15_hermann}, the document $D$ is usually a text document of $n$ words. The query $Q$ is a question about the document and there might be $m$ questions per document. In practice, $m$ is undefined. The cost of current softmax attention mechanisms may prevent real-time question answering from many users.

	\item In information retrieval tasks (such as a search engine), the document $D$ may represent a long sequence (such as a webpage). A query $Q$ could be a single question about a fact implicitly contained in one of these documents $D$. The classic softmax attention mechanism would require scanning all the words of every document $D$ all over again for each new searched query. 

	\item In network architectures with external memory \citep{Graves2014, NIPS2015_5846}, $D$ represents the memory to be queried. Current attention mechanism may limit the size of the memory and the number of queries. It seems particularly important to develop more efficient memory mechanisms. One such possibly would be a memory architecture whose memory size does not scale linearly with the number of facts to be stored. Another one would be a linear size memory but a sublinear query algorithm.

\end{itemize}

More generally, the softmax attention mechanism is prohibitive in large-scale applications which have long sequences ($n >> k$), an extremely high amount $m$ of queries (possibly to be processed in real-time) and strong memory constraints. There is thus a potential interest for developing cheaper attention mechanisms that would satisfy the following properties:

\begin{itemize}
	\item At test time, a computational complexity independent from the document size $n$, by opposition to the $O(n)$ complexity of current attention mechanisms. Such a cheap attention would have very little overhead compared to a recurrent model with no attention (in terms of the sequence size $n$).
	\item At test time, a fixed-size representation of the document, by opposition to the $O(n)$ memory representations of current attention mechanisms.
	\item At training time, if there are $m$ queries per document, an algorithm which does not scale in $O(nm)$ but only in $O(n)$.
\end{itemize}

The linear attention mechanism that we introduce in the next section satisfies these requirements, allowing to potentially tackle problems at a much larger scale. As expected, our early experiments show that these computational gains come at the price of slightly worse accuracy than the softmax attention mechanism, yet definitively better than no attention.

\section{Cheap linear attention mechanism}

\subsection{Definition and complexity}

In this section, we introduce the simplest version of the linear attention mechanism; more sophisticated additions are described in the next section. The linear attention mechanism results from the removal of the softmax, leading to the following linear attention mechanism:

\[
R(D, Q) = H^T H q = C q,
\]

where $C = H^T H$ is a square matrix of dimension $k \times k$. $C$ represents a non-centered covariance matrix of the hidden states, it is computed in $O(n k^2)$ complexity. Most importantly, it depends only on the document $D$ (not on the query $Q$). This implies that if $C$ is computed once, any attention lookup will only cost $O(k^2)$, i.e. with a complexity independent from $n$, the length of the document sequence. For $m$ queries, the resulting attention complexity would be $\boldsymbol{O(m k^2)}$, i.e. a $n$ speedup compared to the classic softmax attention mechanism ($O(m n k^2)$). Furthermore, each document $D$ can be summarized into  the matrix $C$, i.e a fixed-size representation of size $k \times k$ instead of the $k \times n$ matrix of hidden states required by the softmax attention. Note that if $k > n$ there is no memory improvement, in which case it is more suitable to store $H$ rather than the singular matrix $C$ of rank $k$. Notice that $C$ can be seen as the non-centered covariance matrix of the hidden states.

\subsection{Computation of C}

The matrix $C$ is equal to $C = H^T H$. Computing it that way still requires to store all the hidden states $h_{(t)}$ and then perform a huge matrix multiplication. To avoid this $O(n \times k)$ memory footprint at test time, we can notice that

\[
C = H^T H = \sum_{t=1}^n h_{(t)} h_{(t)}^T,
\]

which suggests an iterative way to compute it:

\[
C_{(t+1)} = C_{(t)} + h_{(t+1)} h_{(t+1)}^T,
\] 

and $C = C_{(n)}$. This iterative process avoids storing all the hidden states and the matrix $C$ can eventually be computed using only $O(k^2)$ memory space. 

Although the complexity of computing $C$ is still linear in the size of the sequence $n$, this computation has to be done only a single time per document, which contrasts with the classic attention mechanism, for which we have to scan all over again the document for each new query $Q$.

\subsection{Backpropagation through $C$}

Using the iterative procedure to compute $C$ does not require to store all the intermediate $C_{(t)}$ during backpropagation. The attention lookup process $Cq$ can be written as

\[
Cq = \sum_{t=1}^n h_{(t)} h_{(t)}^T q = \sum_{t=1}^n c_{(t)},
\]

where $c_{(t)} = h_{(t)} h_{(t)}^T q$. Naive automatic differentiation tools may save all the states of the matrix $C$ in the forward pass, which is unnecessary given that the corresponding gradient of the loss $L$ with respect to $h_{(t)}$ can be written as:

\[
\nabla_{h_{(t)}} = q \left(h_{(t)}^T \nabla_{c_{(t)}} \right) + \nabla_{c_{(t)}} \left(h_{(t)}^T q\right),
\]

which shows that it is unnecessary to store the intermediate states $C_{(t)}$.

\subsection{Summary of the computational advantages}

%
%
%


Table~\ref{table:comparison} summarizes the computational and memory benefits of using linear attention mechanisms compared to the original softmax attention. The forward encoding pass is slightly more expensive for the linear attention mechanism because it has to perform an outer product at each timestep to update the matrix $C$.

\begin{table}[!ht]  
  \centering
  \begin{tabular}{|c|c|c|}
    \cline{2-3}
    \multicolumn{1}{c|}{} & Softmax attention & Linear attention \\ \hline 
    a) Query complexity & $O(n k)$   & $O(k^2)$ \\ \hline
    b) Document compression & $n \times k$   & $k \times k$ \\ \hline
    c) Encoding complexity & $O(n k^2 \lambda)$   & $O(n k^2 (\lambda +1))$ \\ \hline
  \end{tabular}
  \captionsetup{width=.8\linewidth}
  \caption{Comparison between the traditional softmax mechanism and the linear mechanism of a) the computational cost of an attention lookup, b) the memory requirements to store an encoded document and c) the computational cost of encoding the document ($\lambda$ is a constant depending on the type of recurrent unit).}
  \label{table:comparison}
\end{table}

\section{Gated linear attention mechanisms}

We can generalize the cheap linear attention described previously by incorporating non-linear functions to update $C_{(t)}$:

\[
C_{(t+1)} = \alpha_{(t)} C_{(t)} + \beta_{(t)} f_{(t)} f_{(t)}^T,
\]

where $\alpha_{(t)}$, $\beta_{(t)}$ and $f_{(t)}$ are (non-linear) functions of $h_{(t+1)}$ and $C_{(t)} f_{(t)}$. Their intended functions are described as follows:

\begin{itemize}
	\item The quantity $C_{(t)} f_{(t)}$ is useful because it measures to some extent how much of $f_{(t)}$ is already contained in $C_{(t)}$. Suppose that $C_{(t)}$ already contains $f_{(t)}$ and only other orthogonal vectors to $f_{(t)}$, then $C_{(t)} f_{(t)} = \lVert f_{(t)} \rVert f_{(t)}$, which gives information on the presence or not of $f_{(t)}$ in the matrix $C_{(t)}$.
	\item $\alpha_{(t)}$ and $\beta_{(t)}$ control to what extent the network remembers about the previous $C_{(t)}$.
	\item $f_{(t)}$ lets the network precisely update certain regions of the matrix $C_{(t)}$. $f_{(t)}$ could be the element-wise product of $h_{(t+1)}$ and a sigmoid whose input is $h_{(t+1)}$.
\end{itemize}

Backpropagation requires to know the intermediate values of $C_{(t)}$ at each timestep. Instead of storing them in the forward pass, which would be prohibitive memory-wise, we can incrementally re-compute each $C_{(t)}$ starting from the final matrix $C = C_{(n)}$ and invert the successive transformations. If we memorize in the forward pass the values of $\alpha_{(t)}$, $\beta_{(t)}$, $f_{(t)}$ and $h_{(t)}$, we can use them to compute $C_{(t)}$ from $C_{(t+1)}$:

\[
C_{(t)} = \frac{C_{(t+1)} - \alpha_{(t)} f_{(t)} f_{(t)}^T}{\beta_{(t)}}.
\]

Theano implementations of this backward pass and code for the experiments are available on our github repository\footnote{\href{https://github.com/adbrebs/efficient_attention}{https://github.com/adbrebs/efficient\_attention}}.

In the experiments below, we use a particular instance of the general model above, which we call \emph{gated linear attention}. It is defined by $\alpha_{(t)} = \beta_{(t)} = 1$ and $f_{(t)} = \sig (W h_{(t+1)} + b) \odot  h_{(t+1)} $, where $\odot$ is the element-wise product. In other words, the network has now the capacity to control the information it adds to the matrix $C$. The full mechanism can be written as:

\[
C_{(t+1)} = C_{(t)} + \left(\sig (W h_{(t+1)} + b) \odot h_{(t+1)}\right) \left(\sig (W h_{(t+1)} + b) \odot h_{(t+1)}\right)^T.
\]

\section{Experiments}

\begin{figure}[!htb]
	\includegraphics[width=\textwidth]{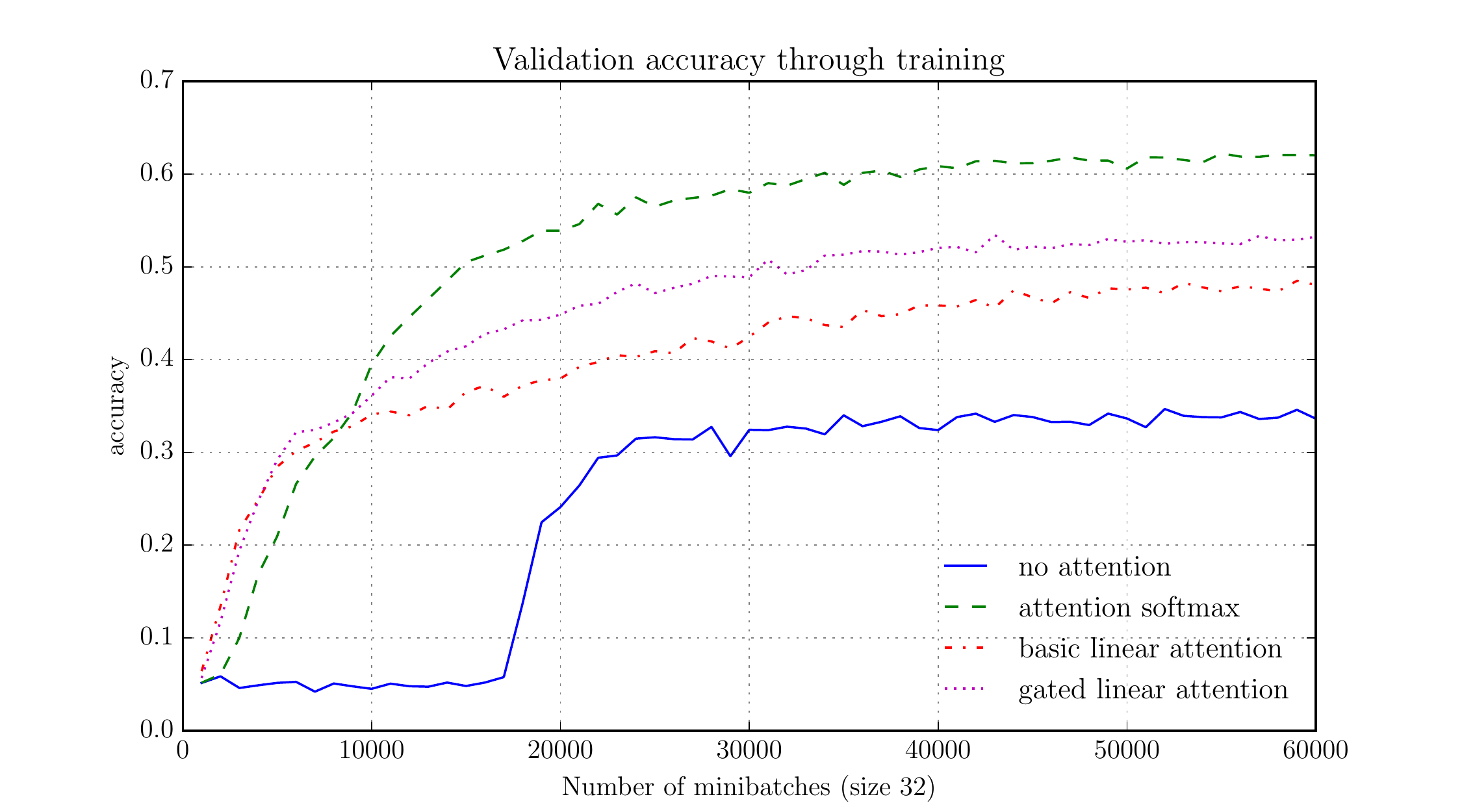}
	\vspace{-1.5em}
	\caption{Comparison of the validation accuracies obtained with different attention mechanisms on the CNN question-answering dataset. We observe a) that as expected the softmax attention mechanism yields the best accuracy, b) that the linear mechanisms are significantly better than no attention, c) that the gated linear attention is significantly better than the basic linear attention, d) that models with attention are faster to converge, probably due the skip connections introduced by the attention mechanism.}
	\label{fig:qa_comparison}
\end{figure}

The cheap linear attention mechanism is designed for large-scale applications with a very large number of queries per document in real-time. Research datasets are not really suitable to highlight their computational efficiency in practice. Therefore, we focus on comparing their accuracy results. 

We evaluated the basic and gated versions of the linear attention mechanism on a question answering task. We used the CNN dataset released by~\citet{nips15_hermann}, which is composed of Cloze style questions on documents with $n=750$ words on average. There are about $m=4$ questions per document. We did not aim to reach state of the art results but simply to compare the different versions of attention. As such, we used a simple architecture, which only requires a few hours to train. We fixed the architecture for all our experiments and the models only differ by their attention part. More precisely, the common architecture is composed of a single-layer GRU network to encode the query and a separate single-layer GRU network to encode the document\footnote{Note that for their baseline model without attention, \citet{nips15_hermann} concatenated the question and the document. Despite improving a lot the performance, this approach does not allow to compute a representation of the document independent of the query (it requires to know the question in advance). Therefore we encoded the query and the document with two independent networks.}. We used ADAM to train our networks. For the two GRU networks, we chose a small hidden size $k=100$ and word embeddings of size 100.

At test time, an optimized implementation should yield a speedup of $\frac{n*k*m}{m*k^2} = \frac{n}{k} \approx 7$ for each attention lookup\footnote{These are the complexity gains for the attention lookups only, we do not consider the forward pass necessary for both softmax and linear attentions.}. However, at this stage, we are more interested in the accuracy results comparison rather than the speed. The speedup would better be illustrated in applications with a (very) large number of queries per document and relatively long documents, but such public datasets are still rare.

\section{Discussion}

Our early experiments on question-answering suggest that linear mechanisms and their gated extensions significantly improve models with no attention. As expected, the accuracy results of softmax attention are better but the gap can be reduced when adding non-linear gates to the basic linear mechanism. We believe that more sophisticated extensions could further improve the results.

In terms of memory, the linear attention mechanisms can be seen as a trade-off between no-attention models and classic softmax models. They compress the document sequence into $k \times k$ representations, which can store more information than the $k$-length vector of the last hidden state of a classic recurrent network, but obviously less than the $n \times k$ stored hidden states of a softmax attention mechanism. This is probably more suitable for tasks with relatively long sequences and an extremely high number of lookups. Nevertheless, for extremely long sequences, we believe that fixed-size representations may not capture enough information and further research should focus on sublinear (maybe $O(\log(n))$ or adaptative, depending on how much information is contained in the sequence) representations.

This $k \times k$ representation can not only store more information than a $k$-length vector but it also acts as skip connections from the past hidden states to the output. As a result, we observed that it can capture longer term dependencies and the training optimization is easier because it is less prone to the vanishing gradient problem.

A potential extension of this cheap mechanism is to interleave the updates of $C_{(t)}$ and $h_{(t)}$ to create a new flavor of recurrent unit, which uses second order information about the past hidden states ($C_{(t)}$ can be seen as a non-centered covariance matrix). The recurrent unit would take as input not only the previous hidden state $h_{(t-1)}$ and the current input $x_{(t)}$ but also the product $C_{(t)} h_{(t)}$ which evaluates to some extent how much of $h_{(t)}$ is already stored in $C_{(t)}$. 

\section{Conclusion}

We introduced a new family of attention mechanisms, called linear attention mechanisms, which, with little computational overhead, yield better and easier to optimize models compared to standard recurrent networks with no attention. Their constant $O(k^2)$ attention lookup complexity and their $O(k^2)$ memory requirements make them very appealing alternatives to build large-scale information retrieval systems, for which the computational costs of traditional softmax attention mechanisms are prohibitive. More precisely, we believe that the linear attention mechanisms would be suitable on large-scale tasks with some of these three properties:
\begin{itemize}
	\item long sequences, long enough so that a recurrent network with no attention is unable to capture long-enough dependencies.
	\item many attention lookups, such that traditional softmax attention mechanisms would be too slow. This is particularly important for real-time systems which have to process extremely large loads of queries simultaneously (for example millions of queries per hour).
	\item a requirement to store documents into fixed-size representations.
\end{itemize}

%
%
%
%
%
%
%
%
%
%
%
%


\bibliography{iclr2016_conference,strings,strings-shorter,ml,zoulou,aigaion-shorter}
\bibliographystyle{iclr2016_conference}

\end{document}